\documentclass[letterpaper, 10 pt, conference]{ieeeconf}  

\IEEEoverridecommandlockouts                              

\usepackage{graphicx} 
\newcommand{\myfigureshrinker}{\vspace{-0.4cm}}

\title{\LARGE \bf
Pain Assessment based on fNIRS using Bidirectional LSTMs
}

\author{Raul Fernandez Rojas$^{1}$, Julio Romero$^{1}$, Jehu Lopez-Aparicio$^{2}$, and Keng-Liang Ou$^{3,4,5}$
\thanks{$^{1}$Human-Centred Technology Research Centre, Faculty of Science and Technology, University of Canberra, ACT, Australia.
        {\tt\small raul.fernandezrojas@canberra.edu.au}}%
\thanks{$^{2}$Faculty of Science, National Autonomous University of Mexico, Mexico City, Mexico.}%
\thanks{$^{3}$Department of Dentistry, Taipei Medical University Hospital, Taipei, Taiwan.}%
\thanks{$^{4}$Department of Dentistry, Taipei Medical University-Shuang Ho Hospital, New Taipei City, Taiwan.}%
\thanks{$^{5}$3D Global Biotech Inc., New Taipei City, Taiwan.}%
\thanks{*This work was partially supported by 3D Global Biotech Inc.}
}

\begin{document}

\maketitle
\thispagestyle{empty}
\pagestyle{empty}

\begin{abstract}
Assessing  pain  in  patients  unable  to  speak  (also called non-verbal patients) is extremely complicated and often is  done  by  clinical  judgement.  However,  this  method  is  not reliable since patients’ vital signs can fluctuate significantly due to other underlying medical conditions. No objective diagnosis test  exists  to  date  that  can  assist  medical  practitioners  in  the diagnosis of pain. In this study we propose the use of functional near-infrared  spectroscopy  (fNIRS)  and  deep  learning  for  the assessment of human pain. The aim of this study is to explore the  use  deep  learning  to  automatically  learn  features  from fNIRS raw data to reduce the level of subjectivity and domain knowledge required in the design of hand-crafted features. Four deep  learning  models  were  evaluated,  multilayer  perceptron (MLP),  forward  and  backward  long  short-term  memory  net-works  (LSTM),  and  bidirectional  LSTM.  The  results  showed that the Bi-LSTM model achieved the highest accuracy ($90.6\%$)and  faster  than  the  other  three  models.  These  results  advance knowledge in pain assessment using neuroimaging as a method of   diagnosis   and   represent   a   step   closer   to   developing   a physiologically based diagnosis of human pain that will benefit vulnerable  populations  who  cannot  self-report  pain.

\end{abstract}

\section{INTRODUCTION}

Assessing pain in patients unable to speak (also called non-verbal patients) is extremely complicated and often is done by clinical judgement. However, this method is not reliable since patients’ vital signs, including blood pressure, heart rate, and respiration, fluctuate significantly due to underlying medical conditions \cite{arbour2010vital}. In addition, assessing pain in non-verbal patients becomes extremely difficult in situations where the patient is in endotracheal intubation, in mechanical ventilation, in comma, under surgery, or with cognitive impairment such as dementia. These situations limit healthcare professionals' ability to appropriately respond to and treat the patient's discomfort \cite{van2017moving}. Untreated pain might lead to health deterioration, unnecessary suffering, delay recovery, worsen cognitive impairments, or increase the risk of death. These medical conditions and the obstacle to evaluate and manage patient’s pain experience in an appropriate manner calls for the need for a reliable and objective pain assessment.


Possible methods to measure and assess pain objectively have been proposed in the literature. In a recent study \cite{sobas2020evaluation}, different pain biomarkers in the saliva of subjects were explored, with concentrations of secretory immunoglobulin A and tumor necrosis factor-alpha presenting acceptable levels of reproducibility in healthy subjects; however, salivary-based biomarker variations are not only due to pain-related information \cite{sobas2020evaluation}. Another example is the use of cerebrospinal fluid which has been reported to identify pain in patients with neuropathic pain and movement disorders in patients with complex regional pain syndrome \cite{jain2010handbook}. For gastrointestinal disorders, the use of pharmacological agents such as fentany and octrotide has showed magnitudes of change in sensory end points, however, there is no clear evidence of an effective biomarker of visceral pain \cite{jain2010handbook}. Another proposed biomarker of pain is the use serum of catecholamine (stress hormone) which reflects sympathetic activation \cite{goldstein2012cerebrospinal}. These examples have shown that biological samples can be used as possible indicators of human pain.

Recently, neuroimaging methods such as, magnetic resonance imaging (MRI), positron-emission
tomography (PET), electroencephalography (EEG), or functional near-infrared spectroscopy (fNIRS),  have gained recognition due to their ability to gain insights into the human brain
(noninvasively) and understand the components involved in pain processing. In this context, fNIRS has demonstrated that external pain stimulation can evoke changes of oxygenation levels in distinctive cortical regions \cite{rojas2016spatiotemporal}. fNIRS measures brain activity by reading cerebral haemodynamics and oxygenation. Specifically, it measures changes in chromophore mobilization, oxygenated haemoglobin (HbO) and deoxygenated haemoglobin (HbR) simultaneously \cite{rojas2017toward}. In addition, fNIRS offers advantages over the other technologies such as, better temporal and spatial resolution, less exposure to ionising radiation, safe to use over long periods and many times, relatively inexpensive, and portable; advantages that make fNIRS a possible candidate for the development of a bedside monitor for the diagnosis of pain.

Partially responsible to the success of fNIRS and other neurimaging methods to assess human pain is the use of machine learning. In pain research, machine learning and fNIRS has successfully been applied for detection and prediction of pain. For instance, Pourshoghi et al., \cite{pourshoghi2016application} used a SVM classifier to classify low pain and high pain from 19 subjects using coefficients from functional data analysis (FDA) as features with $94\%$ accuracy. In another study \cite{rojas2019machine}, Rojas et al., used LDA, KNN and SVM classifiers to explore features in time, frequency, and wavelet domain, showing that SVM achieved $88.41\%$ accuracy to classify heat and cold pain-related data. 

In these examples, machine learning was used to discover a mapping of features from fNIRS data, and then to determine a pain neural representation than can be used to assess or predict new data. However, a limitation of machine learning models is the use of hand-crafted feature extraction, which requires human intervention to design and select the most relevant features that enable the learning models to solve the task at hand. Therefore, deep learning methods that can learn features automatically from fNIRS data are expected to reduce subjectivity in the design of hand-crafted features and improve overall model performance.

This paper contributes to the development of an objective assessment of pain for non-verbal patients. In particular, we explore the use of deep learning methods to reduce the level of subjectivity and domain knowledge to design hand-crafted features, and instead, process directly raw fNIRS data. 
Therefore, the main objective of this study is 
to examine different deep learning models to identify an improved method for the assessment of pain. This study represents a step closer to developing a physiologically-based assessment of human pain that would benefit non-verbal patients.    


\section{METHODOLOGY}

\subsection{Subjects}
Eighteen right-handed volunteers (three females) were considered in the study, with 
mean age $\pm$ standard deviation (31.9 $\pm$ 5.5). The reason to include right-handed participants was to avoid any variation in functional response due to lateralisation of brain function. Written informed consent was obtained from all volunteers. This research study was approved by the Taipei Medical University (TMU) and
full-board review process of the TMU-Joint Institutional Review Board under contract number 201307010.

\subsection{Experimental Procedure}
In this study, thermal pain perceptions were investigated following the standard procedures of the QST protocol \cite{rolke2006quantitative}. The pain experiment involved applying cold and heat to the skin to induce pain. Subjects were exposed to gradually increasing or decreasing temperatures with a thermode and pressed a button when they experienced pain (threshold test) and highest intensity of pain (tolerance test). The experimental protocol was split into two tests, the thermal pain threshold (low pain) and the thermal pain tolerance (high pain), with a 2-min rest between both tests. After an initial 60-s rest, the experiment started with a random selection of three consecutive trials of cold or heat stimulation, with a 60-s rest between tests. Based on the threshold and tolerance of cold and heat stimuli, the fNIRS data were organized into four categories: (1) low-cold (low pain), (2) low-heat (low pain), (3) high-cold (high pain), and (4) high-heat (high pain). These categories were used to label the database for the classification task.

\subsection{fNIRS Equipment}
Cortical haemodynamic data were collected using an optical topography system, Hitachi ETG-4000 (Hitachi Medical Corporation, Japan). The spectrometer system is equipped with a 24-channel cap 
configured in 12 channels per hemisphere. The measurement area was the somatosensory cortex, using the international 10-20 system, probes were centred on the C3 and C4 positions. Only raw Oxyhaemoglobin (Oxy-Hb) was used in the analysis due to its higher signal-to-noise ratio compared to Deoxyhaemoglobin (Deoxy-Hb). The sampling frequency was $10Hz$ and no pre-processing was applied to the fNIRS data. 


\subsection{Deep Learning Models}
In this section the fundamentals of the models are introduced. Three deep learning models are explored, multi-layer perceptron (MLP), long short-term memory (LSTM) networks, and bidirectional LSTM (Bi-LSTM).

\subsubsection{MLP}
They are a computational model that processes information using a number of interconnected nodes. These nodes are grouped into layers and associated through weighted connections. Nodes in each layer apply non-linear operations to project the data into a space where the data becomes linearly separable \cite{ordonez2016deep}. Every neuron (or computational unit) can be defined in terms of the following function: 

$$
    a^{l+1} = \sigma(W^{l}a^{l}+b^{l}), \eqno{(1)}
$$

where $a^{l}$ refers to the activation value for the neurons in the layer $l$, $W$ is the weight matrix, $b{l}$ is the bias of neurons in layer $l$, and $\sigma$ is the activation function. In the topology of MLPs fully-connected layers are employed, in these, each unit in layer $l+1$ is connected with every unit in layer $l$. The MLP model will be used as baseline model and to obtain reference values to compare the LSTM and Bi-LSTM models.

\subsubsection{LSTM}
These networks are a type of recurrent neural network (RNN) and suitable for processing time series data. LSTMs were designed to overcome the problem of long-term dependency in RNN, which made long sequences difficult to learn using RNN due to the vanishing/exploding gradient problem \cite{alhagry2017emotion}. LSTMs use a gated cell to control what information should be remembered and what information should be forgotten, this is achieved by three gates. The first gate is a forget gate to determine the information to be removed from the cell state using a $sigmoid$ layer  

$$
    f_{t} = \sigma (W_{f} \cdot [h_{t-1},x_{t}] + b_{f}),    \eqno{(2)}
$$

The second gate is an input gate that uses a $sigmoid$ layer to determine the values that are updated, and a $tanh$ layer to define new updated values, formally defined as:

$$
    i_{t} = \sigma (W_{f} \cdot [h_{t-1},x_{t}] + b_{i}),   \eqno{(3)}
$$
$$
    \tilde{c_{t}} = \tanh (W_{c} \cdot [h_{t-1},x_{t}] + b_{c}).    \eqno{(4)}
$$

The cell state is then updated by:

$$
    c_{t} = f_{t} * c_{t-1} + i_{t} * \tilde{c_{t}}    \eqno{(5)}
$$

The output gate determines the output in the current stated using the cell state (Equation 5) and a sigmoid layer:

$$
    o_{t} = \sigma (W_{o} \cdot [h_{t-1},x_{t}] + b_{o}),    \eqno{(6)}
$$
$$
    h_{t} = o_{t} * \tanh (c_{t}),    \eqno{(7)}
$$

where $x_{t}$ is the input sequence at time $t$, $h_{t-1}$ refers to the past hidden state and $b_{f}$, $b_{i}$, $b_{c}$, $b_{o}$ are the bias vectors in each layer. 

In this study, the forward LSTM (standard method) and the backward LSTM (reversed input sequences) will be implemented to compare the performance to the Bi-LSTM. 

\subsubsection{Bi-LSTM}
This model is a combination of a forward and a backward LSTM, in other words, at every time step, data are input in both forward and reverse directions at the same time. This model concatenates the outputs of the standard and reverse LSTM by specifying the merge mode in Keras. Bi-LSTMs preserve information from both past and future as data flows in two ways. It is expected that this property will make the Bi-LSTM have higher precision than conventional LSTMs, as LSTMs run inputs in an unidirectional way.    

\subsection{Architectures}
For comparison purposes, all four models share a similar architecture with an input layer, two main hidden layers, and an output layer. In addition, network architectures were determined empirically by multiple experimental adjustments. However, a number of hyper parameters were left constant (default value in Keras) across all three models. This is done with the idea that differences in performance between the models are the result of architectural difference and not due to better optimisation, pre-processing or ad hoc customisation. 

The input layers in all models are fed with raw fNIRS data. Each data input was segmented using a sliding window approach with $50\%$ overlap, with each window containing 300 samples. The input data for the MLP model were in the form of a $m \times n$ matrix, where $m$ is the number of samples and $n$ refers to the number of fNIRS channels. The LSTM models expect the data in the form of $m \times k \times n$ where $k$ refers to the time steps.

The proposed MLP model consists of two fully connected layers. The first layer and second layer have 64 and 32 neurons, respectively, and a rectifier linear activation unit (ReLu) was used in these two layers. A dropout layer with a $50\%$ dropout rate was used to prevent over-fitting \cite{srivastava2014dropout}. Similarly, the architecture of the LSTM networks consists of two LSTM layers. The two LSTM layers consist of 64 neurons and 32 neurons respectively, and both layers with a ReLU activation function. Then a dropout layer with $50\%$ dropout rate after the second LSTM layer.     

Finally, the output layer in all three architectures includes a fully connected layer with four neurons corresponding to the four pain classes in the experiment and Softmax activation for classification. In addition, all models are trained using Adam algorithm to optimise the networks and the categorical cross entropy loss function for the multi-class classification task.


\section{RESULTS}
 In order to evaluate the deep learning models, the dataset was randomly divided into training set ($70\%$) for  and test data ($30\%$). 10-fold cross validation was used on all networks to avoid over-fitting. Training was done for 300 epochs with an early stopping criterion of halting training when there is no decrease in error during the last 50 epochs. The network with the highest accuracy is chosen as the final model. The deep learning models were trained with a learning rate of 0.001, decay factor of 0.9, and batch size of 64. 
 
 The performance on the test set of the MLP, backward LSTM, forward LSTM, and Bi-LSTM are listed in Table \ref{table:results}. In this table, accuracy, sensitivity and specificity were used as the evaluation methods for the test data. Overall, the Bi-LSTM showed the highest accuracy with $90.6\%$, the forward LSTM (standard) with $85.6\%$, and the backwards LSTM with $82.2\%$, and the baseline model (MLP) with $71.7\%$.

\begin{table}[h]
\caption{Comparison of the performance of CNN, forward LSTM (F-LSTM), backward LSTM (B-LSTM), and Bi-LSTM models.}
\label{table:results}
\begin{center}
\begin{tabular}{l c c c}
\hline
\hline
Model   & Accuracy ($\%$) & Sensitivity ($\%$) & Specificity ($\%$) \\
\hline
MLP     & $71.7$ & $57.3$ & $75.4$\\
F-LSTM  & $85.6$ & $77.9$ & $88.2$\\
B-LSTM  & $82.2$ & $73.4$ & $86.6$\\
Bi-LSTM & $90.6$ & $84.6$ & $90.4$\\
\hline
\hline
\end{tabular}
\end{center}
\end{table}


\begin{figure}[b]
      \centering
      \myfigureshrinker
      \setlength{\fboxsep}{0pt}%
      \setlength{\fboxrule}{0pt}%
      \framebox{\includegraphics[scale=0.6]{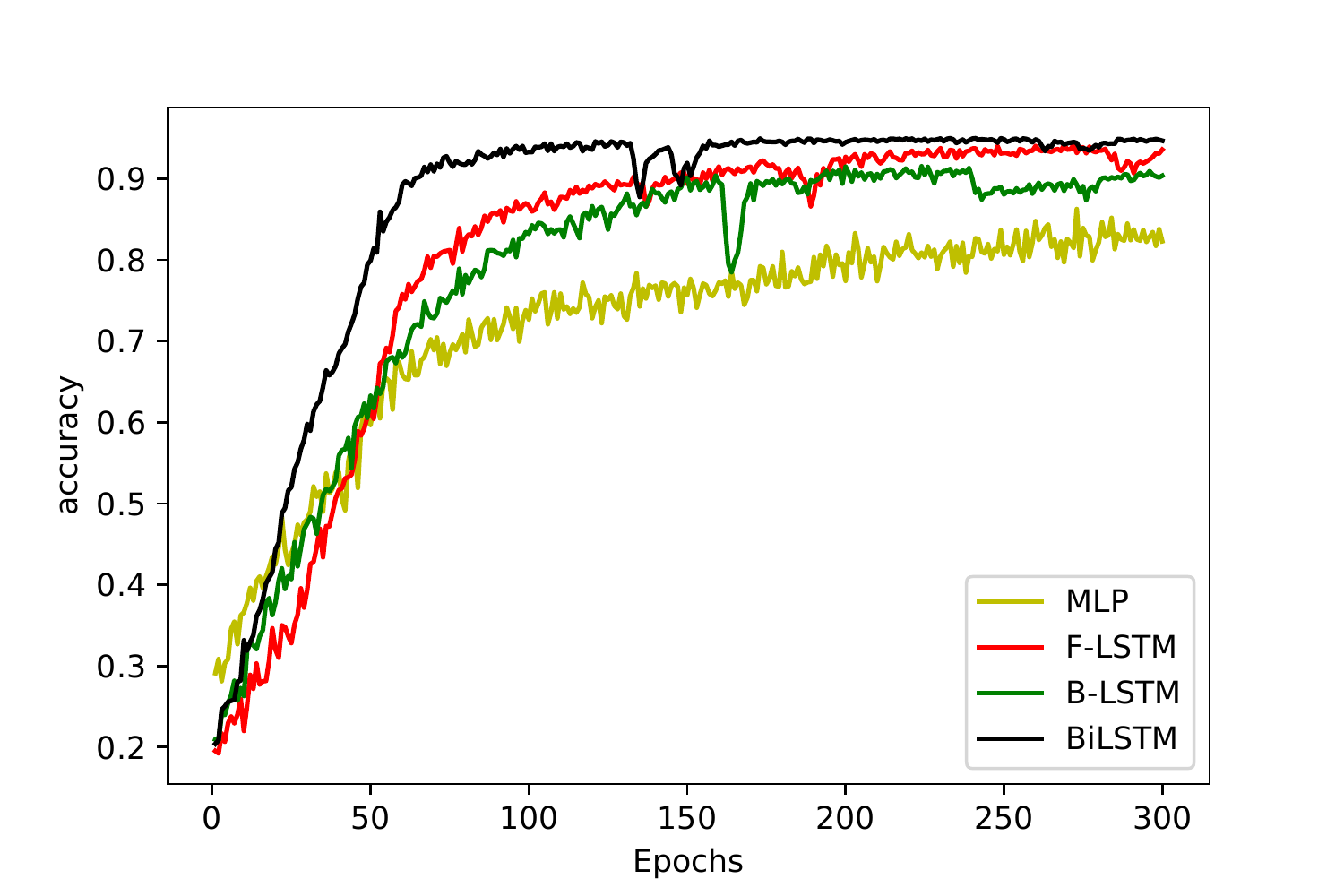}}
      \caption{Average accuracy on the validation set over 300 epochs from the deep learning models.}
      \label{figure:acc}
   \end{figure}

\begin{figure}[thpb]
    \myfigureshrinker
      \centering
      \includegraphics[scale=0.6]{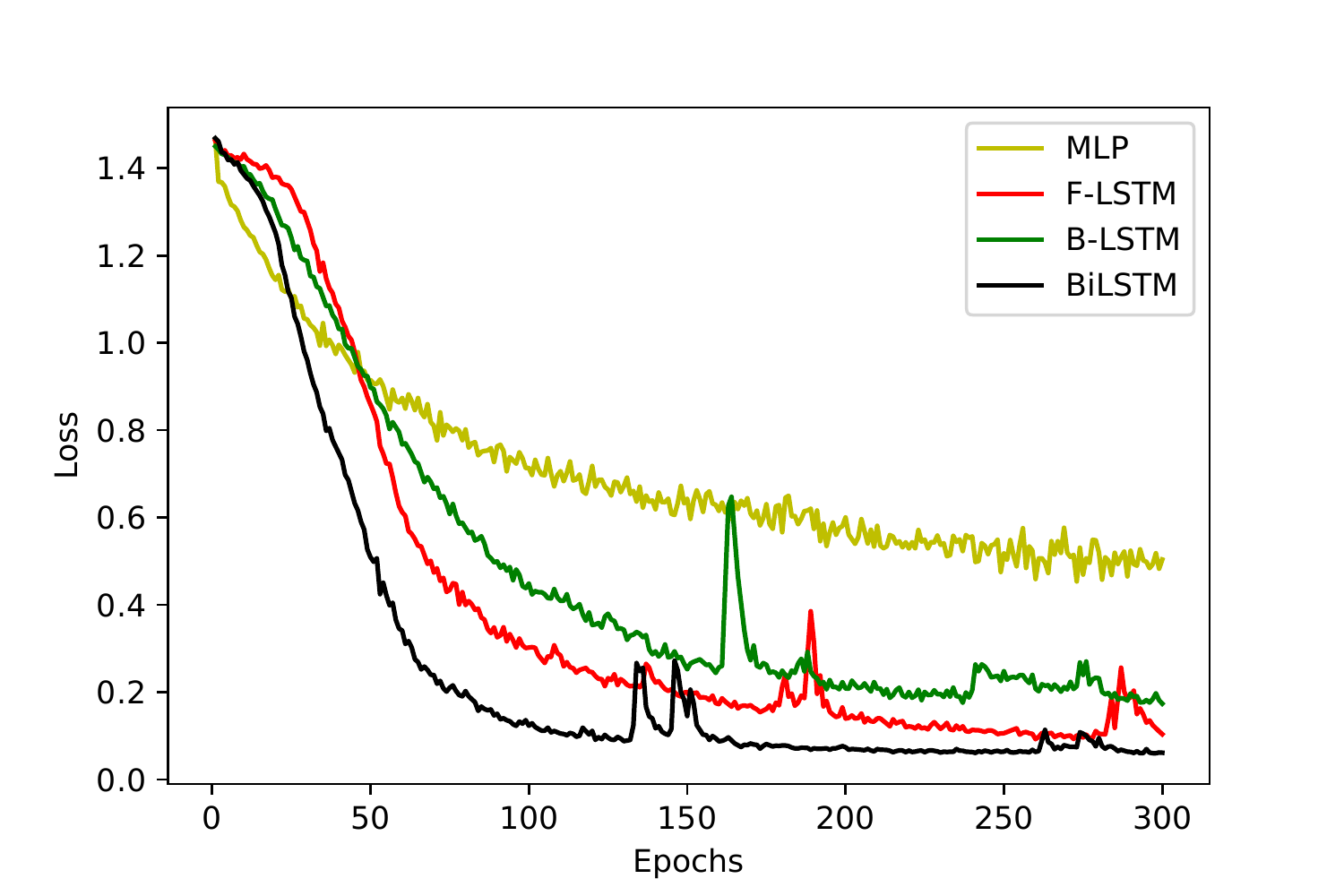}
      \caption{Average loss on the validation set over 300 epochs from the deep learning models.}
      \label{figure:loss}
   \end{figure}

The performance of the training process of all networks with different number of iterations is presented in Figure \ref{figure:acc}. During the training/validation process the Bi-LSTM achieved $93.1\%$ accuracy using around 100 epochs, this was twice faster than the other two LSTMs ($\sim$ 200 epochs) to reach comparable results at $90\%$. This might allow a reduction in the number of epochs required in the training process, which can be important in training larger datasets using similar architectures. Similar results can be observed in Figure \ref{figure:loss}, where the Bi-LSTM reached a more stable loss faster that the other three networks.

These results showed that the deep learning models exhibited comparable results with previous work. For example, in \cite{pourshoghi2016application} the authors achieved $94\%$ accuracy using a SVM with a total of 120 features generated by functional data analysis decomposition. Similarly, in \cite{rojas2019machine} $88.41\%$ accuracy was obtained with a Gaussian SVM with 69 features. An advantage of the present study is that the deep learning models are fed with raw fNIRS data, while these previous studies used pre-processed fNIRS data and hand-crafted features.


It is also important to highlight that this study has some limitations. Direct comparisons with other published studies in the literature are difficult due to the use of different experimental conditions, different neuroimaging methods (e.g., fMRI, PET), number of subjects, or learning models. Another limitation in this study is the lack of control for any skin blood flow contributions and intracerebral haemodynamics to the fNIRS signals. Recent studies have highlighted the issue that fNIRS signals encompass not only haemodynamic fluctuations due to neurovascular coupling but also due to skin blood flow and task-related systemic activity of cortex \cite{rojas2017physiological}. However, it was expected that the use of deep learning models will learn the neural representation of pain from raw fNIRS data.


\section{CONCLUSIONS}

In this study, we explored the use of deep learning models to learn features automatically from fNIRS raw data and also to identify an improved method of the assessment of pain for non-verbal patients. The experimental results showed that the combination of backward and forward information in the architecture of the  Bi-LSTM model achieve better results than the one-directional LSTMs (fordward/backward). In addition, the use of deep learning models did not require complex feature extraction procedures to obtain satisfactory results, this also reduces the level of subjectivity in the design of hand-crafted features. These results advance knowledge in pain assessment using neuroimaging as a method of diagnosis and represent a step closer to developing a physiologically based diagnosis of human pain that would benefit vulnerable population who cannot self-report pain.

Future work will focus on using different sensor modalities (e.g., heart rate, respiration, galvanic skin response, electroencephalogram). A multimodal approach that combines data from several sensors together can be more robust and/or more comprehensive for real-time assessment of human pain. This multimodal approach will help identify possible relationships between different sensor modalities, or complement each other in case of artifacts or signal quality issues. In addition, the use of multimodal data fusion techniques will help unlock the potential of multiple sensors in our future work.

\addtolength{\textheight}{-12cm}   


\bibliographystyle{plain}
\bibliography{IEEEfull}

\end{document}